\documentclass{article}

\usepackage[margin=1in]{geometry}
\usepackage{setspace}
\usepackage[T1]{fontenc}
\usepackage{lmodern}
\usepackage{microtype}

\usepackage{amsmath,amssymb,amsfonts}
\usepackage{bm}

\usepackage{graphicx}
\usepackage[svgnames]{xcolor}
\usepackage{booktabs}
\usepackage{multirow}
\usepackage{colortbl}

\usepackage[font=small,labelfont=bf,skip=4pt]{caption}
\usepackage[colorlinks=true,linkcolor=blue,citecolor=blue,urlcolor=blue]{hyperref}

\usepackage[sort&compress,numbers]{natbib}

\usepackage{enumitem}
\usepackage{adjustbox}
\usepackage{subcaption}
\usepackage{float}
\usepackage{placeins}

\title{\textbf{CIGPO: Contextual Information-Gain Policy Optimization \\
for Multi-Turn Evidence-Reading LLM Agents}}

\author{Hao Dou\\[4pt]
\small Harbin Institute of Technology}

\date{}

\begin{document}
\maketitle

\begin{abstract}
Training multi-turn evidence-reading agents with outcome-only reinforcement
learning is unstable because intermediate turns receive little direct credit.
In HotpotQA experiments with Qwen2.5-3B-Instruct, GRPO initially improves
(standard F1 0.430) but subsequently collapses to 100\% format-violating
outputs. Training-log diagnosis reveals a zero-advantage lock-in mechanism: all
sampled trajectories receive the minimum format penalty ($-2.0$), group-relative
advantages vanish, and the policy-gradient loss becomes zero---an optimization
deadlock. We propose a variance-injection strategy: by assigning per-turn rewards
to intermediate evidence-reading turns, we prevent the group reward distribution
from collapsing to a single value---preserving the variation that GRPO's
group-relative advantage requires. \textbf{C}ontextual
\textbf{I}nformation-\textbf{G}ain \textbf{P}olicy \textbf{O}ptimization (CIGPO)
implements this strategy using the marginal increase in the frozen reference
model's log-likelihood of the ground-truth answer as the per-turn signal. With
separate normalization of IG and F1 rewards and an IG-weight curriculum, CIGPO
reaches a standard F1 of 0.518 on HotpotQA at the 3B scale (from 0.252 base;
+105\%), compared with 0.430 for the best GRPO checkpoint and 0.000 for the
final GRPO checkpoint. CIGPO maintains meaningful reward variance and avoids
zero-advantage lock-in throughout training. These results identify
reward-variance collapse as a concrete failure mode of outcome-only GRPO and
show that turn-level IG rewards can prevent it in this HotpotQA setting.
\end{abstract}

\section{Introduction}

Large language models (LLMs) are increasingly deployed as autonomous agents that
interact with external tools and knowledge sources over multiple turns
\cite{yao2022react,schick2023toolformer,wang2024planning}. A canonical task in this
paradigm is \emph{multi-turn evidence reading}: given a question, the agent decides
which documents to read, retrieves evidence, and synthesizes an answer across
multiple rounds of retrieval and reasoning.

Reinforcement learning (RL) is a natural framework for training such agents.
Recent advances in RL for LLMs, particularly Group Relative Policy Optimization
(GRPO)~\cite{shao2024deepseekmath}, have shown impressive results in math reasoning
and code generation by optimizing a sparse outcome reward (e.g., final answer
correctness). However, we observe a critical stability issue when applying GRPO to
multi-turn evidence reading: GRPO initially improves (standard F1 reaches 0.430 at
step 50 on Qwen2.5-3B-Instruct) but subsequently collapses, with the model
degenerating to 100\% format-violating outputs by step 150. Training logs at the point of collapse show that critic score mean converges to
exactly $-2.000$, the zero-advantage group ratio reaches $1.0$, and the
policy-gradient loss becomes exactly $0.000$. The resulting entropy loss spikes
to $\sim$4809, consistent with garbled high-entropy output distributions from
which the model cannot recover without an external learning signal.

We identify the root cause as an \textbf{optimization deadlock} arising from the
credit assignment problem in multi-turn RL: when only the final turn receives a
non-zero reward, intermediate turns lack a learning signal. The policy drifts
toward degenerate behaviors---such as omitting required XML tags or generating
garbled output---that, while scoring the minimum format penalty ($-2.0$), produce
zero advantage when all group members also produce violations. Without variation
in group rewards, GRPO provides no meaningful gradient, and the policy collapses
into a deadlock from which it cannot recover.

To address this, we adopt a \textbf{variance-injection} principle: the key
requirement is to ensure that the reward distribution across GRPO group members
retains non-zero variance, making group-relative advantages less likely to vanish. As
one concrete implementation, we propose \textbf{Contextual Information-Gain
Policy Optimization} (CIGPO). CIGPO is a lightweight extension of GRPO that
injects variance at intermediate turns via \emph{reference-based contextual
information gain}. Information gain
quantifies how much each evidence-reading turn improves the reference model's
confidence in the correct answer:

\begin{equation}
r_t^{\mathrm{IG}} = \log p_{\mathrm{ref}}(y^* \mid q, e_{\leq t}) -
                    \log p_{\mathrm{ref}}(y^* \mid q, e_{<t})
\label{eq:ig_def}
\end{equation}

where $p_{\mathrm{ref}}$ is the frozen reference model (the same model used for KL
regularization in GRPO), $y^*$ is the ground-truth answer, $q$ is the question,
and $e_{\leq t}$ denotes evidence read up to turn $t$.

CIGPO adapts information-gain-based credit assignment
\cite{wang2024igpo} to a controlled evidence-reading environment.
Unlike search-based settings where the model formulates free-form queries, CIGPO
operates in a closed evidence pool: at each turn the agent selects and reads
pre-indexed evidence blocks via structured \texttt{Read[block\_id]} actions.
The IG reward is assigned as a sparse turn-level signal---placed on the last token
of each intermediate turn---while the standard F1 reward is assigned to the last
token of the final turn.

Our contributions are:
\begin{itemize}[nosep]
    \item We diagnose the training instability of outcome-only GRPO in multi-turn
          evidence reading as an optimization deadlock, showing that GRPO initially
          improves but then collapses to 100\% format violations via zero-advantage
          lock-in in our Qwen2.5-3B HotpotQA setting.
    \item We propose a variance-injection approach to multi-turn credit
          assignment, and implement it as CIGPO---which adapts reference-based
          contextual information gain to evidence-reading turns---demonstrating
          that it stabilizes training at the 3B scale, improving standard F1
          from 0.252 to 0.518 on HotpotQA.
    \item Through analysis of per-turn IG values and training-log metrics, we
          provide evidence that IG rewards are correlated with successful evidence
          use (mean cumulative IG of 4.06 for correct vs.\ 2.10 for wrong
          trajectories) and that CIGPO maintains meaningful reward variance
          throughout training, preventing the zero-advantage lock-in observed in
          GRPO.
\end{itemize}

We restrict our claims to the Qwen2.5-3B HotpotQA setting under constrained
hardware (2$\times$24GB GPUs). The goal is to document a specific failure
mode---zero-advantage variance collapse---and to test whether a turn-level
reward signal can prevent it in this setting.

\section{Related Work}

\textbf{RL for LLM Agents.}
Reinforcement learning has become a standard approach for aligning and improving
LLMs. Proximal Policy Optimization (PPO)~\cite{schulman2017ppo} and its variants
have been widely adopted for RLHF~\cite{ouyang2022instructgpt,bai2022constitutional}.
Direct Preference Optimization (DPO)~\cite{rafailov2023dpo} offers a simpler
alternative by optimizing preferences directly, though it does not natively
handle multi-turn credit assignment. More recently,
GRPO~\cite{shao2024deepseekmath} eliminates the value network by
using group-level relative advantages, as in DeepSeekMath and DeepSeek-R1
\cite{guo2025deepseekr1}. These methods have proven effective for math and code
generation where a single-turn response suffices. Our work extends GRPO to
multi-turn settings by adding per-turn credit assignment.

\textbf{Multi-Turn Agent Training.}
Several works explore RL training for multi-turn agent interactions.
ReAct~\cite{yao2022react} and Toolformer~\cite{schick2023toolformer} enable LLMs
to use tools in a multi-turn fashion but do not use RL for optimization.
Agent Q~\cite{putta2024agentq} and StreamBench~\cite{zhang2024streambench}
apply RL to multi-turn agent tasks but rely on value-based methods or Monte Carlo
tree search. {ReST-MCTS*}~\cite{zhang2024restmcts}
uses tree search with process rewards to provide turn-level signals for search
agents.

\textbf{Information Gain as Reward.}
Information gain has a long history in feature selection and active learning
\cite{settles2009active}. In the context of LLM agents, IGPO~\cite{wang2024igpo}
proposes using information gain to reward search queries in web-based information-seeking
agents. Our work, CIGPO, adapts this idea to a different setting: controlled
evidence reading from a closed document pool rather than open-ended web search.
Table~\ref{tab:igpo_cigpo} summarizes the key differences.

\begin{table}[t]
\centering
\caption{Comparison of IGPO and CIGPO. CIGPO adapts IG-based credit assignment to
controlled evidence-reading environments.}
\label{tab:igpo_cigpo}
\begin{tabular}{lcc}
\toprule
\textbf{Aspect} & \textbf{IGPO} & \textbf{CIGPO} \\
\midrule
Action type & Search query generation & Evidence block selection \\
Action space & Open-ended web search & Closed evidence pool \\
Environment & Web / search engine & Local pre-indexed documents \\
Action format & Free-form search query & Read[evidence\_id] \\
IG computation & IG over search context & IG over evidence context \\
Base optimizer & Various RL algorithms & GRPO \\
Domain & Multi-turn search agents & Multi-turn evidence-reading agents \\
\bottomrule
\end{tabular}
\end{table}

\textbf{Credit Assignment in RL.}
The credit assignment problem is fundamental in RL~\cite{minsky1961steps,sutton2018reinforcement}.
In sequence-level RL for LLMs, prior work has explored token-level reward
redistribution using learned critics~\cite{wu2023finegrained} and process reward
models~\cite{lightman2023process}. CIGPO offers a simpler alternative: the IG
signal provides immediate, per-turn feedback using only a frozen reference model,
without requiring a separately trained reward model.

\section{Preliminaries}

\subsection{Group Relative Policy Optimization (GRPO)}

GRPO~\cite{shao2024deepseekmath} optimizes a policy $\pi_\theta$ by comparing
responses within a group. For each prompt $x$, a group of $N$ responses
$\{o_i\}_{i=1}^N$ is sampled from the current policy. Each response receives a
reward $r_i$, and advantages are computed using group-level normalization:

\begin{equation}
A_i = \frac{r_i - \mu_{\text{group}}}{\sigma_{\text{group}}}
\label{eq:grpo_adv}
\end{equation}

where $\mu_{\text{group}}$ and $\sigma_{\text{group}}$ are the mean and standard
deviation of rewards within the group. The policy is then updated via a clipped
surrogate objective:

\begin{equation}
\mathcal{L}_{\text{GRPO}} = \mathbb{E}\left[
\min\left(\rho_i A_i,\; \text{clip}(\rho_i, 1-\epsilon, 1+\epsilon) A_i\right)
- \beta \, D_{\text{KL}}(\pi_\theta \| \pi_{\text{ref}})
\right]
\label{eq:grpo_loss}
\end{equation}

where $\rho_i = \pi_\theta(o_i|x) / \pi_{\text{old}}(o_i|x)$ is the importance
weight, $\epsilon$ is the clipping threshold, and $\beta$ controls the KL penalty
toward a reference policy.

\subsection{Multi-Turn Evidence Reading}

We formalize multi-turn evidence reading as a sequential decision process. Given a
question $q$ and a set of candidate evidence documents $\mathcal{E} = \{e_1, \ldots, e_K\}$,
the agent interacts for up to $T$ turns. At each turn $t$, the agent:
\begin{enumerate}[nosep]
    \item Reasons about what information is needed (in a \texttt{<think>} block)
    \item Calls a tool to read a document (via \texttt{<tool\_call>} with a
          structured \texttt{Read[block\_id]} action)
    \item Receives evidence content (in a \texttt{<tool\_response>} block)
\end{enumerate}
After the final turn (or earlier if the agent chooses), the agent produces a final
answer in an \texttt{<answer>} tag. The answer is scored against the ground truth
using token-level F1.

In standard GRPO-based training, only the final turn's answer token receives a
non-zero reward. All intermediate turns---the reasoning and tool-calling
decisions---receive zero reward.

\section{CIGPO Method}

\begin{figure}[t]
\centering
\includegraphics[width=\textwidth]{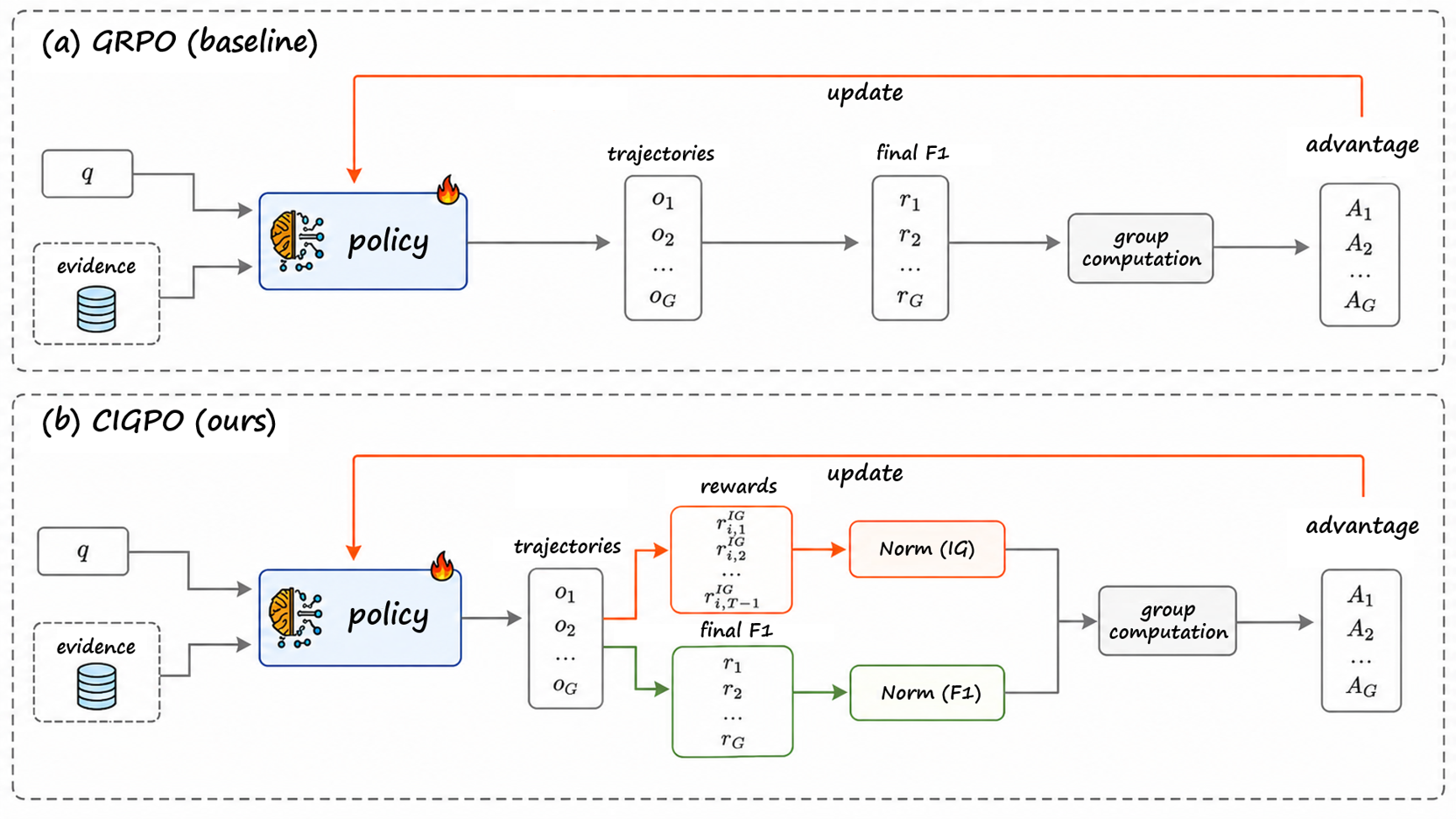}
\caption{Overview of the CIGPO training framework. Given a question and a pool of
pre-indexed evidence documents, the policy model (trained via GRPO with IG-augmented
rewards) selects and reads evidence blocks across multiple turns. At each intermediate
turn $t$, the contextual information gain $\Delta_t = \log p_{\mathrm{ref}}(y^* \mid q, e_{\leq t}) -
\log p_{\mathrm{ref}}(y^* \mid q, e_{<t})$ is computed using the frozen reference model
and assigned as a per-turn reward. The final turn receives the standard F1 reward
against the ground-truth answer. Separate group normalization is applied to IG and
F1 rewards before advantage computation.}
\label{fig:arch}
\end{figure}

CIGPO extends GRPO with three components designed for multi-turn credit
assignment. The unifying principle is variance injection: GRPO's group-relative
advantage requires reward variation within each group to produce a meaningful
gradient. When all trajectories in a group receive identical rewards---as when
every sample produces a format violation---advantages vanish and the policy
receives no learning signal. CIGPO prevents this by ensuring intermediate
evidence-reading turns carry non-zero, varied rewards.
Figure~\ref{fig:arch} contrasts the reward structure of GRPO and CIGPO.

\subsection{Per-Turn Information Gain Reward}

The core of CIGPO is rewarding each evidence-reading turn with its information gain.
For turn $t$ (where $1 \leq t < T$), we define:

\begin{equation}
r_t^{\text{IG}} = \log p_{\pi_{\text{ref}}}(y^* \mid q, e_1, \ldots, e_t) -
                 \log p_{\pi_{\text{ref}}}(y^* \mid q, e_1, \ldots, e_{t-1})
\label{eq:ig_reward}
\end{equation}

where $p_{\pi_{\text{ref}}}$ is the probability under the frozen reference model.
We use the \emph{reference} model (rather than the current policy) for three reasons:
(1) it avoids non-stationary reward drift as the policy updates; (2) it is
consistent with the reference model already used for KL regularization in GRPO;
and (3) it makes IG computation stable and reproducible, as the reference model is
frozen throughout training.

The final turn $T$ receives the standard F1 reward:
\begin{equation}
r_T^{\text{F1}} = \text{F1}(\text{extract\_answer}(o_T), y^*)
\label{eq:f1_reward}
\end{equation}

\textbf{Reward placement.} IG rewards are assigned as sparse turn-level signals:
the IG value is placed on the last token of each intermediate turn (terminal token
reward placement), and the F1 reward
is placed on the last token of the final turn. All other tokens receive zero reward.
No turn-level discounted returns are used. This creates a sparse but turn-granular
reward structure that provides feedback at each decision point without requiring
dense per-token credit assignment.

\subsection{Prealigned Vectorized GT LogProb}

Computing IG naively requires $T$ separate forward passes of the reference model
(one per turn prefix). We accelerate this via \emph{prealigned vectorized
computation}: all turn prompts are padded to a uniform length and concatenated into
a single batch. The reference model computes $\log p(y^* \mid \cdot)$ for all $T$
turns in one batched forward pass, achieving a $\sim$3$\times$ speedup over
sequential computation.

\subsection{Separate GRPO Group Normalization}

A critical design choice is \emph{how} IG and F1 rewards are normalized before
computing advantages. GRPO normalizes rewards within a group using
Equation~\ref{eq:grpo_adv}. However, IG and F1 operate on fundamentally different
scales: IG values are in the nat scale (mean $\approx$~8, range $[-17, +29]$ in
practice), while F1 is bounded in $[0, 1]$ (or receives a format penalty of $-2.0$).

We use \textbf{separate normalization}: IG and F1 rewards are normalized
independently within each group. For a group of $N$ trajectories, let
$\mathcal{R}_{\text{IG}}$ be the set of all intermediate-turn IG rewards and
$\mathcal{R}_{\text{F1}}$ be the set of final-turn rewards. Advantages are computed as:

\begin{equation}
A_t^{\text{IG}} = \frac{r_t^{\text{IG}} - \mu_{\mathcal{R}_{\text{IG}}}}
                      {\sigma_{\mathcal{R}_{\text{IG}}}},\qquad
A_T^{\text{F1}} = \frac{r_T^{\text{F1}} - \mu_{\mathcal{R}_{\text{F1}}}}
                      {\sigma_{\mathcal{R}_{\text{F1}}}}
\label{eq:separate_norm}
\end{equation}

We apply a wide safety cap of $\pm 50.0$ to IG values to prevent numerical
instability while preserving the natural variance that GRPO normalization requires.
We discuss the sensitivity to this choice in Section~\ref{sec:limitations}.

\subsection{Curriculum Schedule}

We employ a linear curriculum that interpolates the IG reward weight from
$\lambda_{\text{IG}}^{\text{init}} = 0.1$ to $\lambda_{\text{IG}}^{\text{final}} = 0.3$
over $S = 200$ training steps:

\begin{equation}
\lambda_{\text{IG}}(s) = \lambda_{\text{IG}}^{\text{init}} +
    \frac{s}{S} \big(\lambda_{\text{IG}}^{\text{final}} - \lambda_{\text{IG}}^{\text{init}}\big)
\label{eq:curriculum}
\end{equation}

The combined per-token reward is $r_t = \lambda_{\text{IG}}(s) \cdot r_t^{\text{IG}}$
for intermediate turns and $r_T = r_T^{\text{F1}}$ for the final turn.
The curriculum starts with a low IG weight to let the model first learn format
compliance, then gradually increases the influence of evidence-reading quality.
The F1 weight is fixed at 1.0 throughout training.

\section{Experimental Setup}

\textbf{Model and Infrastructure.}
We use Qwen2.5-3B-Instruct (3.09B parameters) from the Qwen2.5-Instruct family~\cite{qwen2025qwen2.5}. Training runs
on 2$\times$ NVIDIA GPUs
(24GB each) with tensor model parallelism of size 2. We use FSDP for distributed
training and vLLM~\cite{kwon2023vllm} for efficient rollout generation. Each
training run completes 200 steps (approximately 2.5 hours on our hardware).

\textbf{Data.}
We use HotpotQA~\cite{yang2018hotpotqa} for multi-hop question answering.
The training set contains 1,000 examples; the test set contains 1,000 examples
for final evaluation. Evidence documents are pre-indexed locally.

\textbf{Training Configuration.}
Key hyperparameters are listed in Table~\ref{tab:config}.
The rollout uses a temperature of 0.6 and a maximum of 3 turns per trajectory.
GRPO group size $N=2$, PPO mini-batch size of 2, and KL coefficient of 0.10.
Due to hardware constraints (2$\times$24GB GPUs), group size and batch size are
limited to 2; the effect of larger group sizes remains untested.
Format violations (missing or broken XML tags, or answers without evidence
retrieval) receive a penalty of $-2.0$.

\begin{table}[t]
\centering
\caption{Training hyperparameters.}
\label{tab:config}
\begin{tabular}{lr}
\toprule
\textbf{Hyperparameter} & \textbf{Value} \\
\midrule
Model & Qwen2.5-3B-Instruct (3.09B) \\
Max turns & 3 \\
Rollout temperature & 0.6 \\
GRPO group size $N$ & 2 \\
PPO mini-batch size & 2 \\
Learning rate & $1 \times 10^{-6}$ \\
KL coefficient $\beta$ & 0.10 \\
Entropy coefficient & 0.01 \\
IG curriculum range & $0.1 \rightarrow 0.3$ \\
IG safety clip & $\pm 50.0$ \\
Normalization mode & separate \\
Format penalty & $-2.0$ \\
Max sequence length & 3072 tokens \\
Training steps & 200 \\
\midrule
GRPO Baseline & IG weight = 0 (disabled) \\
\bottomrule
\end{tabular}
\end{table}

\textbf{Evaluation Metrics.}
We evaluate checkpoints every 50 steps on the full 1,000-example test set.
We report two primary metrics:
\begin{itemize}[nosep]
    \item \textbf{Standard F1}: Token-level F1 score with format-violating outputs
          scored as 0. This metric is always in $[0, 1]$.
    \item \textbf{Format-penalized reward}: Training reward including the $-2.0$
          penalty for format violations. This can be negative and is reported
          separately from F1 to avoid confusion.
\end{itemize}
We also report exact match (EM), with format violations scored as 0 (standard EM
in $[0,1]$), separately from the format-penalized EM reward.\footnote{In the
original evaluation, format violations received EM = $-2.0$. We rename this column
to ``Format-penalized EM reward'' and report standard EM separately.}

\section{Results}

\subsection{Best-Checkpoint vs Final-Checkpoint Comparison}

Table~\ref{tab:best_final} provides a fair comparison using both best and final
checkpoints.

\begin{table}[t]
\centering
\caption{Best-checkpoint and final-checkpoint comparison. Standard F1 excludes
format penalty (violations scored as 0). Format-penalized reward includes the
$-2.0$ penalty. Format violation rate is computed from the evaluation logs.}
\label{tab:best_final}
\begin{tabular}{lcccc}
\toprule
\textbf{Model} & \textbf{Standard F1} & \textbf{Penalized} & \textbf{FmtViol} & \textbf{Step} \\
 & \textbf{(no penalty)} & \textbf{Reward} & \textbf{Rate \%} & \\
\midrule
Base Qwen2.5-3B & 0.252 & $-$0.930 & 59.1 & -- \\
\midrule
GRPO (best) & 0.430 & 0.008 & 21.1 & Step 50 \\
GRPO (final) & 0.000 & $-$2.000 & 100.0 & Step 200 \\
\midrule
CIGPO (final) & \textbf{0.518} & 0.272 & 12.3 & Step 200 \\
\bottomrule
\end{tabular}
\end{table}

\textbf{GRPO improves then collapses.}
GRPO reaches a standard F1 of 0.430 at step 50, but format violations rise from
21.1\% (step 50) to 90.5\% (step 100), reaching 100\% by step 150. The final
checkpoint (step 200) produces exclusively format-violating outputs (standard
F1 = 0.000).

\textbf{CIGPO stabilizes and continues improving.}
CIGPO's standard F1 increases from 0.429 (step 50) to 0.518 (step 200; +21\%
relative over step 50, +105\% over the base model's 0.252). Format violations
decrease from 31.5\% to 12.3\%. We note that standard EM peaks at step 100
(0.242) and then declines to 0.150 at step 200, suggesting that continued
training broadens partial answer recall at the expense of exact-match precision
(see Section~\ref{sec:f1_em}).

\subsection{Training Dynamics}

\begin{figure}[t]
\centering
\includegraphics[width=\textwidth]{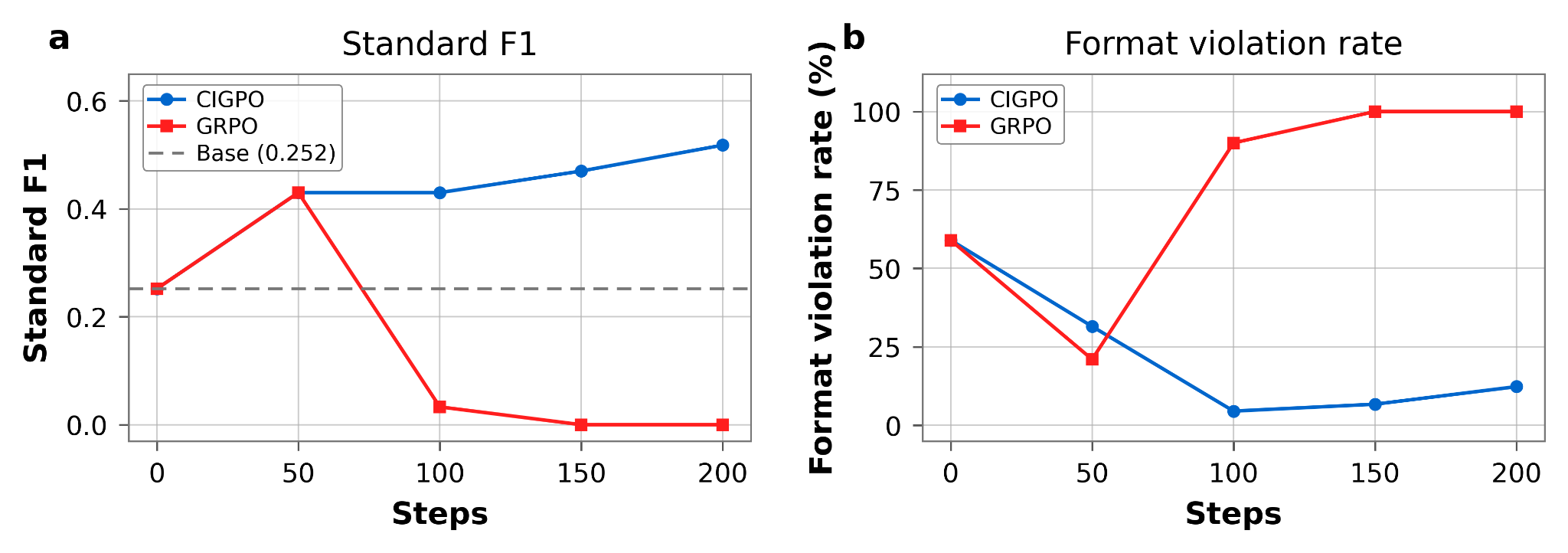}
\caption{Training dynamics. \textbf{Left:} Standard F1 (format violations scored as 0).
CIGPO improves from 0.429 to 0.518; GRPO initially reaches 0.430 at step 50 but
collapses to 0.000. \textbf{Right:} Format violation rate. GRPO violations rise
from 21\% to 100\%; CIGPO violations decrease from 32\% to 12\%.}
\label{fig:dynamics}
\end{figure}

Figure~\ref{fig:dynamics} shows the standard F1 and format violation rate across
checkpoints.

\FloatBarrier

\subsection{Full Benchmark Results}

Table~\ref{tab:full_benchmark} presents the complete
per-checkpoint results.

\begin{table}[H]
\centering
\caption{Full test set results across all checkpoints (HotpotQA, 1,000 test examples).
Standard F1: format violations $\rightarrow$ 0. Penalized reward: violations
$\rightarrow$ $-2.0$. Standard EM: violations $\rightarrow$ 0.}
\label{tab:full_benchmark}
\begin{tabular}{lccc}
\toprule
\textbf{Checkpoint} & \textbf{Standard F1} & \textbf{Penalized Reward} & \textbf{Standard EM} \\
\midrule
Base Qwen2.5-3B & 0.252 & $-$0.930 & 0.011 \\
\midrule
CIGPO Step 50  & 0.429 & $-$0.201 & 0.000 \\
GRPO Step 50   & 0.430 & 0.008 & 0.000 \\
\midrule
CIGPO Step 100 & 0.432 & \textbf{0.344} & \textbf{0.242} \\
GRPO Step 100  & 0.033 & $-$1.781 & 0.000 \\
\midrule
CIGPO Step 150 & 0.471 & 0.339 & 0.217 \\
GRPO Step 150  & 0.000 & $-$2.000 & 0.000 \\
\midrule
CIGPO Step 200 & \textbf{0.518} & 0.272 & 0.150 \\
GRPO Step 200  & 0.000 & $-$2.000 & 0.000 \\
\bottomrule
\end{tabular}
\end{table}

\subsection{Collapse Diagnosis}

Table~\ref{tab:collapse} provides a detailed breakdown of format violations across
checkpoints, computed from the evaluation prediction logs.

\begin{table}[t]
\centering
\caption{Collapse diagnosis across checkpoints. Values computed from 1,000-sample
evaluation JSONL files. ``AnsWoRetr'' = answer without evidence retrieval.
``HasRetr'' = trajectory contains at least one evidence retrieval.}
\label{tab:collapse}
\small
\setlength{\tabcolsep}{3.5pt}
\begin{tabular}{llccccccc}
\toprule
\textbf{Method} & \textbf{Step} & \textbf{Valid\%} & \textbf{FmtViol\%} & \textbf{NoAns\%} & \textbf{Unclosed\%} & \textbf{HasRetr\%} & \textbf{HasAns\%} & \textbf{F1$>$0.5\%} \\
\midrule
Base & 0 & 40.9 & 59.1 & 58.7 & 25.9 & 92.0 & 41.3 & 25.2 \\
\midrule
GRPO & 50 & 78.9 & 21.1 & 19.0 & 0.9 & 97.4 & 81.0 & 42.2 \\
GRPO & 100 & 9.5 & 90.5 & 88.2 & 60.7 & 96.2 & 11.8 & 3.1 \\
GRPO & 150 & 0.0 & 100.0 & 100.0 & 100.0 & 0.0 & 0.0 & 0.0 \\
GRPO & 200 & 0.0 & 100.0 & 100.0 & 100.0 & 0.0 & 0.0 & 0.0 \\
\midrule
CIGPO & 50 & 68.5 & 31.5 & 31.5 & 0.4 & 100.0 & 68.5 & 42.1 \\
CIGPO & 100 & 95.6 & 4.4 & 2.8 & 0.5 & 98.5 & 97.2 & 41.8 \\
CIGPO & 150 & 93.4 & 6.6 & 4.8 & 0.9 & 99.1 & 95.2 & 45.9 \\
CIGPO & 200 & 87.7 & 12.3 & 12.1 & 0.8 & 99.6 & 87.9 & \textbf{51.3} \\
\bottomrule
\end{tabular}
\end{table}

At GRPO steps 150--200, all outputs have unclosed XML tags and missing
\texttt{<answer>} tags, with zero evidence retrieval---the model produces
unstructured token sequences. In contrast, CIGPO maintains high retrieval rates
(99--100\%) and valid format rates (88--96\%) throughout training, with correct
answers (F1 $>$ 0.5) increasing from 42.1\% to 51.3\%.

To further characterize the GRPO collapse from the optimizer's perspective, we
examine the training log metrics at step 200. Table~\ref{tab:training_log_diag}
provides the key metrics that distinguish collapsed GRPO runs from stable
CIGPO training, based on metrics directly recorded during training.

\begin{table}[t]
\centering
\caption{Training-log diagnosis of GRPO collapse at step 200. All values are taken
directly from the GRPO 3B training log at step 200.}
\label{tab:training_log_diag}
\begin{tabular}{lr>{\raggedright\arraybackslash}p{0.35\textwidth}}
\toprule
\textbf{Metric} & \textbf{GRPO Step 200} & \textbf{Interpretation} \\
\midrule
critic/score/mean & $-2.000$ & All sampled trajectories receive the minimum format penalty \\
critic/rewards/mean & $-2.000$ & Reward distribution collapses to the minimum score \\
critic/advantages/mean & $0.000$ & No group-relative advantage remains \\
critic/returns/mean & $0.000$ & Return signal vanishes \\
actor/pg\_loss & $0.000$ & Policy-gradient update vanishes \\
actor/entropy\_loss & $\approx 4809$ & Output distribution becomes highly unstable / high entropy \\
training/zero\_advantage\_group\_ratio & $1.000$ & All groups have zero advantage \\
curriculum/ig\_weight & $0.000$ & GRPO has no intermediate IG signal \\
\bottomrule
\end{tabular}
\end{table}

The training logs (Table~\ref{tab:training_log_diag}) confirm that the GRPO failure corresponds to a
zero-advantage lock-in. In the step-200 log, both \texttt{critic/score/mean} and
\texttt{critic/rewards/mean} are $-2.000$, indicating that all sampled trajectories
receive the minimum format penalty. Since group members receive identical rewards,
the group-relative advantage vanishes, as reflected by
\texttt{critic/advantages/mean} $= 0.000$ and
\texttt{zero\_advantage\_group\_ratio} $= 1.000$. The policy-gradient loss also
becomes $0.000$, indicating that the policy receives no effective learning signal
to recover from format collapse.

\subsection{Per-Turn Information Gain Analysis}

To verify that IG rewards are meaningful, we analyze cumulative IG values across
all CIGPO trajectories, grouped by outcome. Table~\ref{tab:ig_analysis} shows
the results.

\begin{table}[t]
\centering
\caption{Cumulative IG by outcome group (all CIGPO checkpoints, 3,972 parsed trajectories).
Correct = standard F1 $>$ 0.5; Partial = $0 <$ F1 $\leq$ 0.5; Wrong = F1 = 0
(valid format); FormatViol = format violation.}
\label{tab:ig_analysis}
\begin{tabular}{lcccc}
\toprule
\textbf{Outcome} & \textbf{N} & \textbf{Mean Cum. IG} & \textbf{Mean Turns} & \textbf{Mean F1} \\
\midrule
Correct (F1 $>$ 0.5) & 1811 & 4.06 & 1.5 & 0.942 \\
Partial (0 $<$ F1 $\leq$ 0.5) & 390 & 3.28 & 1.4 & 0.368 \\
Wrong (F1 = 0, valid) & 1251 & 2.10 & 1.3 & 0.000 \\
Format violation & 520 & 2.97 & 2.0 & 0.000 \\
\bottomrule
\end{tabular}
\end{table}

Trajectories that produce correct answers have the highest cumulative IG (4.06),
while valid wrong trajectories have the lowest IG (2.10). Interestingly,
format-violating trajectories still exhibit non-trivial cumulative IG (2.97),
suggesting that evidence acquisition and final-format compliance are partially
decoupled. This supports the usefulness of IG for evidence
acquisition but also motivates future validity gating to prevent malformed
trajectories from contributing noisy IG rewards. We caution that correlation does
not imply causation---higher IG may reflect easier questions rather than better
evidence-reading strategy.
\section{Discussion}
\label{sec:discussion}

\subsection{Why Does GRPO Collapse in Multi-Turn Settings?}

The GRPO collapse observed in our HotpotQA setting is not merely a degradation in
answer accuracy, but an optimization deadlock. Once format-violating outputs
dominate all samples in a GRPO group, all trajectories receive the same minimum
penalty of $-2.0$. The reward standard deviation collapses, the group-relative
advantage becomes zero, and the policy-gradient loss vanishes---a deadlock from
which the model cannot recover. This is observed
directly in the GRPO step-200 training logs (Table~\ref{tab:training_log_diag}):
score/reward means are $-2.000$, the zero-advantage group ratio is $1.000$, and
actor/pg\_loss is $0.000$. The entropy loss exceeds 4800,
consistent with garbled high-entropy outputs that have lost all task structure.
The collapse is self-reinforcing: once format violations
dominate a group, the policy has no effective learning signal to recover.

CIGPO mitigates this deadlock by supplying sparse turn-level IG rewards at
intermediate evidence-reading turns. Even when final-answer rewards become
homogeneous, variation in per-turn IG preserves relative advantage signals
and reinforces useful evidence acquisition. A single trajectory with relevant
evidence reading produces a positive IG-based advantage for its intermediate
turns, maintaining a gradient toward effective evidence use.

At a conceptual level, CIGPO can be understood as a
\textbf{variance-injection mechanism}. By assigning turn-level IG rewards with
separate normalization, it ensures that the reward distribution across group
members retains non-zero variance even when final-answer rewards are homogeneous.
Whether the injected signal is IG, a learned process reward, or a simpler
heuristic may be secondary---the essential requirement is that intermediate
turns receive a reward signal with sufficient variation to prevent advantage
collapse.

\subsection{What Component of CIGPO Matters Most?}

A limitation of the current study is that we do not isolate which of CIGPO's
three components---the IG reward signal, separate per-metric normalization, or
the IG-weight curriculum---is primarily responsible for stabilizing training.
Under the variance-injection view, these components play complementary roles:
the IG reward supplies a non-constant intermediate-turn signal, separate
normalization keeps this signal from being washed out by the differently scaled
F1 reward, and the curriculum introduces the auxiliary signal gradually after
the model has begun to learn format compliance. The IG signal itself provides
turn-level variation that may prevent the reward distribution from collapsing
even when final-answer rewards are homogeneous. Controlled ablations (IG-only without curriculum, joint
normalization with curriculum, curriculum-only with a constant auxiliary reward)
are needed to disentangle the individual contributions. We leave such ablation
studies to future work, noting that preliminary experiments with aggressive IG
clipping ($\pm 0.5$) and joint normalization collapsed IG variance to near zero,
suggesting that normalization strategy is at minimum a necessary condition for
stability.

\subsection{F1--EM Divergence}
\label{sec:f1_em}

An interesting pattern in the CIGPO results (Table~\ref{tab:full_benchmark}) is
that standard F1 and standard EM move in opposite directions late in training:
F1 rises from 0.432 (step 100) to 0.518 (step 200), while EM declines from 0.242
to 0.150 over the same period. This divergence suggests that continued training
under the combined IG + F1 reward improves the model's ability to retrieve
partially correct tokens (boosting F1) while reducing its tendency to produce
exact string matches (lowering EM). One possible explanation is that the IG reward
guides the model toward reading relevant evidence, which increases token overlap
with the ground truth, but the F1-based final reward does not strongly penalize
paraphrased or approximate answers---so the model learns to generate correct
content without reproducing the exact ground-truth span. This trade-off between
recall (F1) and precision (EM) warrants further investigation in future work.

\subsection{Limitations}
\label{sec:limitations}

We acknowledge several limitations of the current study:

\begin{itemize}[nosep]
    \item \textbf{Single dataset.} Only HotpotQA is used. Generalization to other
          multi-hop QA datasets (2WikiMultihop, MuSiQue), retrieval-augmented
          generation, or web search agent tasks remains future work.
    \item \textbf{No statistical testing.} Results are reported as point estimates
          from a single training run per configuration. Confidence intervals,
          error bars, and significance tests are not reported due to the
          computational cost of multiple independent training runs; the reported
          patterns should be interpreted as suggestive rather than confirmatory.
    \item \textbf{Scale sensitivity.} Although CIGPO stabilizes training in our
          main Qwen2.5-3B HotpotQA setting, preliminary 1.5B runs were unstable for
          both GRPO and CIGPO. This suggests that smaller models may lack sufficient
          capacity to reliably follow the evidence-reading protocol and benefit
          from contextual IG rewards. We therefore restrict our main claims to the
          3B setting and leave systematic scaling studies to future work. Larger
          models (7B--70B) may also exhibit different stability characteristics and
          may be less prone to collapse even without IG.
    \item \textbf{Small group size.} GRPO group size is 2 due to GPU memory
          constraints. With $N=2$, group-relative advantages are binary ($\pm 1$
          when rewards differ, zero when they match), making GRPO inherently more
          brittle than with larger groups where advantage magnitudes can reflect
          reward magnitudes. The zero-advantage deadlock observed here may be
          partially an artifact of this small group size: any pair of identical
          rewards (e.g., two format violations) produces zero advantage, whereas
          with $N \geq 4$, a minority of valid trajectories can still generate
          non-zero advantages. Larger group sizes could affect both the collapse
          dynamics and the effectiveness of separate normalization.
    \item \textbf{IG clipping and normalization sensitivity.} In preliminary
          experiments, aggressive IG clipping ($\pm 0.5$) with joint IG--F1
          normalization collapsed IG variance to near zero ($\sigma \approx 0$),
          producing only 2\% correct outputs. A wide safety clip ($\pm 50.0$) with
          separate per-metric normalization was necessary for stable training;
          isolating the individual contribution of clipping threshold and
          normalization mode remains future work.
    \item \textbf{IG requires ground-truth answers.} IG computation depends on
          $y^*$ and is only available during training. At inference, the model
          relies on the learned policy without IG rewards.
    \item \textbf{Reference-model dependence.} Reference-based IG depends on model
          calibration and may be noisy for questions with long or ambiguous answers.
          The quality of the IG signal is bounded by the reference model's ability
          to assign meaningful probabilities to the ground-truth answer.
    \item \textbf{No validity gating.} The current implementation does not gate
          invalid or malformed evidence-reading turns from IG normalization.
          The non-trivial cumulative IG observed for format-violating trajectories
          in Table~\ref{tab:ig_analysis} suggests that future validity gating may
          improve the signal-to-noise ratio by excluding malformed evidence-reading
          turns from IG normalization.
\end{itemize}

\subsection{Future Work}

Scaling CIGPO to larger models (7B--70B) and more complex agent tasks would test
the generality of variance-injection-based credit assignment. Systematic scaling
studies across model sizes (1.5B--70B) are needed to map out the relationship
between model capacity and variance-injection effectiveness. Incorporating validity gating to filter malformed turns from normalization
could improve the signal-to-noise ratio (see Table~\ref{tab:ig_analysis}).
Evidence-level hit-rate analysis against HotpotQA supporting facts and
investigation of the F1--EM trade-off observed in late-stage training are left
for future work. Future work could also isolate the effects of clipping threshold
and normalization mode through controlled single-variable experiments. Integrating
CIGPO with process reward models~\cite{lightman2023process} could combine learned
per-step verification with the principled IG signal.

\section{Conclusion}

We identified a zero-advantage variance collapse in GRPO-based multi-turn
evidence-reading training---all trajectories receive the same $-2.0$ penalty,
advantages vanish, and pg\_loss reaches zero---and proposed a variance-injection
strategy to prevent it. CIGPO implements this strategy via sparse turn-level
contextual information-gain rewards, reaching a standard F1 of 0.518 (from
0.252 base; +0.266). Correct
trajectories exhibit substantially higher cumulative IG (4.06 vs.\ 2.10 for
wrong), consistent with IG reflecting successful evidence use. Our claims are
restricted to the Qwen2.5-3B HotpotQA setting. Preliminary 1.5B experiments
suggest that both GRPO and CIGPO can become unstable at smaller model scales,
highlighting model capacity as an important factor for future work. Key open
questions---disentangling IG, normalization, and curriculum contributions;
testing at larger group sizes; and evaluating on diverse datasets---are left for
future work. The main takeaway is that, for small-group GRPO in multi-turn agent
training, maintaining reward variance at intermediate turns can be as important
as improving the final outcome reward.

\clearpage
\appendix

\section{Qualitative Examples}
\label{sec:qualitative}

We present three representative examples from the evaluation prediction logs.

\subsection{GRPO Collapsed Example (Step 150)}

\begin{itemize}[nosep]
    \item \textbf{Question}: Billy Preston is an American college basketball player
          for a team that competes in what?
    \item \textbf{F1 (no fmt penalty)}: 0.0
    \item \textbf{Format-penalized reward}: $-2.0$
    \item \textbf{Analysis}: The GRPO model at step 150 produces garbled output
          with no \texttt{<answer>} tag, no coherent evidence retrieval, and
          unclosed XML tags. The output consists of random token sequences in
          multiple languages/scripts, representing complete format collapse.
          This qualitative collapse matches the training-log diagnosis
          (Table~\ref{tab:training_log_diag}): at this stage, all sampled
          trajectories receive the $-2.0$ format penalty, the zero-advantage
          group ratio reaches $1.0$, and the policy-gradient loss is $0.0$.
          The base model and CIGPO both answer this question correctly (answer:
          ``NCAA'' / ``Kansas Jayhawks'' / ``NCAA Division I'').
\end{itemize}

\noindent\textbf{Trajectory (first 300 chars)}:
\begin{quote}\small\begin{verbatim}
<think>.capitalizeia small Zheng ...
[...garbled output across multiple scripts...]
\end{verbatim}\end{quote}

\subsection{CIGPO Success Example (Step 200)}

\begin{itemize}[nosep]
    \item \textbf{Question}: The Cineplex Odeon Theatres are now operated as which
          American movie theater chain majority-owned by Chinese conglomerate
          Dalian Wanda Group?
    \item \textbf{Evidence IDs read}: [4]
    \item \textbf{Predicted answer}: AMC Theatres
    \item \textbf{F1 (no fmt penalty)}: 1.0
    \item \textbf{Analysis}: CIGPO correctly identifies the relevant evidence block,
          reads it, and extracts the correct answer. The trajectory follows the
          intended format: think $\rightarrow$ tool\_call $\rightarrow$ tool\_response
          $\rightarrow$ think $\rightarrow$ answer.
\end{itemize}

\noindent\textbf{Trajectory (truncated)}:
\begin{quote}\small\begin{verbatim}
<think>From the question, we need to find out what company owns
Cineplex Odeon Theatres... The evidence ID [4] Cineplex Odeon
Corporation could be helpful...</think>
<tool_call>
{"name": "read", "arguments": {"block_id": 4}}
</tool_call>
[...tool_response with evidence text...]
<think>The evidence shows Cineplex Odeon Corporation was acquired
by AMC Theatres, which is majority-owned by Wanda Group.</think>
<answer>AMC Theatres</answer>
\end{verbatim}\end{quote}

\subsection{CIGPO Failure Example (Step 200, Valid Format)}

\begin{itemize}[nosep]
    \item \textbf{Question}: What government position was held by the woman who
          portrayed Shirley in the movie ``The Story of Us''?
    \item \textbf{Evidence IDs read}: [1, 1] (re-read same evidence)
    \item \textbf{Predicted answer}: Actress
    \item \textbf{F1 (no fmt penalty)}: 0.0
    \item \textbf{Analysis}: CIGPO maintains valid format and reads evidence, but
          reads the same evidence block twice (wasting a turn) and produces an
          incorrect answer. The model correctly identifies the actress but fails
          to find her government position. This represents a typical CIGPO failure
          mode: valid format and reasonable evidence selection, but incomplete or
          incorrect multi-hop reasoning.
\end{itemize}

\noindent\textbf{Trajectory (truncated)}:
\begin{quote}\small\begin{verbatim}
<think>I need to find who portrayed Shirley in "The Story of Us"
and what government position she held. Let me check evidence [1].</think>
<tool_call>
{"name": "read", "arguments": {"block_id": 1}}
</tool_call>
[...tool_response with actress info...]
<think>The evidence shows the actress. Let me re-read for
more details.</think>
<tool_call>
{"name": "read", "arguments": {"block_id": 1}}
</tool_call>
[...same tool_response, noting repeat...]
<think>The evidence only mentions her as an actress, not
government position.</think>
<answer>Actress</answer>
\end{verbatim}\end{quote}

\clearpage

\section*{Acknowledgments}

This work was supported by the Harbin Institute of Technology.

\bibliographystyle{plainnat}

\end{document}